\documentclass[lettersize,journal]{IEEEtran}
\usepackage{amsmath,amsfonts}
\usepackage{algorithm}
\usepackage{array}
\usepackage[caption=false,font=normalsize,labelfont=sf,textfont=sf]{subfig}
\usepackage{textcomp}
\usepackage{stfloats}
\usepackage{url}
\usepackage{verbatim}
\usepackage{graphicx}
\usepackage{cite}
\usepackage{mathrsfs}
\usepackage{xcolor}
\usepackage{textcomp}
\usepackage{manyfoot}
\usepackage{booktabs}
\usepackage{algpseudocode}
\usepackage{listings}
\usepackage{pdfpages} 
\usepackage{tabularx}
\usepackage{adjustbox}
\usepackage[pdfpagelabels]{hyperref}
\usepackage{amsmath}
\usepackage[nopatch]{microtype}
\usepackage{booktabs}
\usepackage{url}
\usepackage{graphicx}
\usepackage{multirow}
\usepackage{amsmath,amssymb,amsfonts}
\usepackage{amsthm}
\newcommand{\redtext}[1]{\textcolor{black}{#1}}
\newcommand{\bluetext}[1]{\textcolor{black}{#1}}

\usepackage{xr-hyper}
\begin{document}
% \addbibresource{example.bib}
\title{A Hybrid Framework for Song Lyric Annotation \\ Based on Human-LLM Alignment}

\author{
Rashini Liyanarachchi, 
Frank Tran, 
Md Mahmudul Hasan, 
Aditya Joshi, 
Erik Meijering
\\ \medskip
School of Computer Science and Engineering, University of New South Wales, Sydney, Australia
        % <-this % stops a space
}


\maketitle

\begin{abstract}
Emotion recognition of song lyrics is a challenging task since lyrics may not necessarily align with the overall emotion of a song. As a result, lyrics annotation remains largely underexplored. Drawing inspiration from research in large language model (LLM) assisted annotation, we examine the alignment between humans and LLMs for annotation of lyrics by creating a new sentence-level dataset of lyrics. Our observations highlight the subjectivity of the task and the inherent challenges. Following this, we present a hybrid annotation \redtext{framework\footnote{\url{https://github.com/Rashy98/LyricAnnotator-emotion}}} that optimizes human and LLM annotation by predicting potential misalignment in annotation.
\end{abstract}


\section{Introduction}\label{sec1}

Emotion plays a central role in how we experience music \cite{robinson2005deeper}. While much research in music emotion recognition (MER) has focused on audio features such as tempo, mode, and timbre, textual modality such as lyrics is often either treated as secondary or simply assigned the same emotional label derived from listening to the full song \cite{liyanarachchi2025survey}. Past research overlooks the fact that lyrics carry unique, often evolving emotional meaning that may not align with the music alone. Consequently, most contemporary MER approaches assign emotion labels to a song as a whole, combining the perceived emotions from both audio and lyrics without disentangling their individual contributions.  This oversimplification ignores how lyrics can independently influence emotion perception, even shifting emotional tone line by line, thereby making emotion annotation of lyrics a temporally dynamic and semantically complex task.

However, a growing body of recent research challenges the notion that lyrics are secondary in emotional perception. For instance, \cite{intro_1} found that lyrics significantly shape listeners’ emotional responses, suggesting they should not be treated as interchangeable with audio-based emotion cues. Similarly, \cite{intro_2} emphasized the multifaceted psychological mechanisms that drive emotional reactions to music, with lyrics contributing through narrative and semantic content. However, annotating lyric emotions is inherently difficult due to subjectivity, cultural variation, and linguistic nuances, and is further complicated by the fact that the process is often labor-intensive and time-consuming. 

\cite{intro_3} also demonstrated how subjective and sociocultural biases such as inter-annotator disagreement, individual interpretation, and cultural background, can affect emotion-related tasks in textual domains, complicating the consistency and comparability of annotations. These challenges, spanning cognitive subjectivity, cultural influences, linguistic ambiguity, and annotation scalability, highlight the need for alternative or complementary approaches to emotion annotation. \bluetext{Therefore}, we investigate the research question:
% To address these challenges, We investigate the research question: 
\emph{How well do large language models (LLMs) align with human annotation of emotion in lyrics and, based on this, how can LLMs and humans complement each other in the annotation process?}

We \bluetext{focus on} annotation alignment between humans and LLMs \bluetext{\textit{at the sentence level}}, rather than at the song or verse level, to better capture the temporal and semantic evolution of emotions within lyrics. Sentence-level annotation provides a more precise unit of analysis, aligning with the dynamic, context-sensitive nature of emotional expression in language, an aspect often obscured when assigning a single emotional label to an entire song. 

Our approach to address the research question is motivated by key obstacles in the annotation process: subjectivity, cultural and interpretive variability, semantic ambiguity, and the high cost of manual annotation at scale. 

Building on this, we introduce a hybrid annotation \redtext{framework} that \bluetext{integrates} human judgment with LLM outputs, aiming to capitalize on the complementary strengths of both. Specifically, we explore methods such as weighted aggregation of annotations and predictive modeling to determine the optimal source of annotation for each instance dynamically. The overarching objective is to develop a robust and reliable framework for lyric emotion annotation, where robustness refers to consistency across diverse linguistic and cultural contexts, and reliability is operationalized in terms of inter-annotator agreement and semantic coherence. This framework is intended to facilitate more precise and scalable emotion labeling, thereby supporting the development of more accurate multimodal MER \bluetext{systems. By leveraging} LLMs in combination with human annotators, we aim to explore whether they can (1) increase objectivity and consistency between annotators, (2) reduce susceptibility to cultural bias, (3) capture nuanced emotional cues at a fine-grained level, and (4) significantly reduce the manual effort required for large-scale annotation.

The rest of the paper is organized as follows. Section~\ref{LR} outlines related work. Section~\ref{methods} compares the ability to analyze emotions of humans and LLMs, covering dataset curation, annotation procedures, analyses, and cost considerations. Section~\ref{Hyb_Meth} introduces a hybrid annotation \redtext{framework} combining both sources, including weighted aggregation and annotation source prediction. \redtext{Section~\ref{sec:discussion} discusses the limitations and implications of our study.} Finally, our main findings and suggestions for future work are summarized in Section~\ref{Conc}.

\section{Related Work} \label{LR}
We first outline \bluetext{past work} in dataset collection and annotation, highlighting differences in emotional frameworks and granularity. We then summarize state-of-the-art modeling approaches, from traditional text-based methods to more advanced and multimodal techniques.

\subsection{Lyrics Dataset Collection and Creation}

The development of lyrics-based emotion recognition systems critically depends on the availability and construction of appropriate datasets. Various studies have approached this challenge by collecting, annotating, and curating lyrics datasets from different sources, with diverse emotional taxonomies, granularity levels, and languages.

\cite{An_et_al} focused on collecting lyrics via web scraping (e.g.\ from Baidu Music) and annotating them with emotion categories derived from listener tags. Their dataset of 3,552 songs was filtered into three broad categories---contentment, depression, and exuberance---based on Thayer's two-dimensional model \cite{thayers}. To enhance label quality, ambiguous and overlapping tags were excluded. In contrast, \cite{Choi_et_al} compiled lyrics from Billboard charts and Genius.com, applying the NRC Emotion Lexicon \cite{mohammad2013nrc} to map words in lyrics to eight emotion categories. This lexicon-based approach offered scalable emotion quantification without the need for manual annotation.

Building on the idea of symbolic annotation, \cite{sulun_et_al} introduced Emotion4MIDI, a large-scale symbolic music dataset with annotations adapted from the EMOPIA dataset \cite{emopia}. While not focused exclusively on lyrics, it exemplifies the integration of affective labels such as anger, happiness, sadness, and tenderness, using GoEmotions \cite{goemotion} categories to standardize emotion representation.

To facilitate sentence-level modeling, \cite{Revathy_et_al} curated the Lyric-FM dataset, where each lyric line is annotated with continuous valence and arousal scores. This enables fine-grained learning of emotional nuance across time. The dataset supports supervised training of transformer-based models on sentence-level input, advancing beyond document-level representations.

Multilingual and regional considerations have also shaped recent dataset efforts. \cite{agrawal_et_al} introduced a multilingual lyrics dataset incorporating English, Spanish, and Hindi songs. They utilized translated annotations and XLNet models to accommodate language variability, showing enhanced robustness across linguistic domains. Similarly, \cite{shanker_et_al} developed a Telugu song lyrics dataset annotated in the valence-arousal (VA) space. They leveraged native speaker judgments and applied the \href{https://huggingface.co/docs/transformers/en/model_doc/xlm-roberta}{XLM-RoBERTa} model for domain adaptation, contributing to low-resource language representation in music emotion tasks.

In summary, lyrics emotion datasets vary along several dimensions: annotation type (categorical versus continuous), granularity (song-level versus sentence-level), emotional taxonomy (discrete categories, VA space, or lexicon-based tags), and linguistic diversity (monolingual versus multilingual). These datasets provide the foundation for building increasingly expressive, context-aware models of lyrical emotion, and reflect a growing emphasis on dataset quality, representativeness, and annotation methodology.

\begin{table*}[!t]
    \caption{Comparative overview of lyrics-based music emotion recognition studies.}
    \centering
    \resizebox{\textwidth}{!}{%
    \begin{tabular}{>{\raggedright}p{2cm}>{\raggedright}p{2.5cm}>{\raggedright}p{2cm}>{\raggedright}p{2cm}>{\raggedright}p{3cm}>{\raggedright}p{3cm}>{\raggedright\arraybackslash}p{3.5cm}}
    \toprule
    \textbf{Reference} & \textbf{Dataset Source} & \textbf{Emotion Labels} & \textbf{Annotation Level} & \textbf{Methodology} & \textbf{Key Features/Tools} & \textbf{Notable Aspects} \\
    \midrule
    \cite{laurier_et_al} & Various public datasets & 4 discrete emotions & Song-level & Similarity-based, LSA, LMD + SVM & Lucene, SVM, tf.idf, LMD term selection & Highest accuracy with LMD (80.7\%); multimodal fusion benefit \\ 
    \midrule
    \cite{dakshina_} & Million Song Dataset subset & 8 discrete emotions & Song-level & Latent Dirichlet Allocation (LDA) & Probabilistic topic modeling & Captures probabilistic emotion distributions; 72\% accuracy \\ 
    \midrule
   \cite{An_et_al} & Baidu Music (web crawl) & 15 discrete emotions (filtered to 3) & Song-level & Naive Bayes classifier & Scrapy for crawling, manual filtering & Simplified categories based on Thayer's model; label consistency focus \\ 
    \midrule
   \cite{Choi_et_al} & Billboard Charts, Genius.com & 6 discrete emotions & Song-level & Lexicon-based sentiment analysis & JSOUP parser, NLP preprocessing & Emotional distance quantification; scalable emotion scoring \\ 
    \midrule
    \cite{agrawal_et_al} & Multilingual lyrics datasets & Multiple discrete emotions & Song-level & XLNet transformer & XLNet pretraining, multilingual corpora & Multilingual lyric modeling; better contextual capture \\ 
    \midrule
   \cite{ara_} & Various Datasets & Multiple emotions & Song-level & Text processing & ANEW corpus &  Comprehensive review of emotion analysis in lyrics \\
    \midrule
     \cite{edmonds_} & Dance music excerpts & Multiple emotions & Line-level & BERT-based multi-label classification & BERT model & Captures emotion co-occurrence and ambiguity \\ 
     \midrule
    \cite{shanker_et_al} & Telugu language songs & Valence-Arousal (Russell’s) & Song-level & XLM-RoBERTa fine-tuning & XLM-RoBERTa model & Regional language lyric annotation and recognition \\ 
    \midrule
     \cite{sulun_et_al} & EMOPIA symbolic music dataset & 27 classes (GoEmotions) & Piece-level & Deep learning (transformers) & Large symbolic dataset & Large-scale symbolic music emotion benchmark \\ 
    \midrule
    \cite{Revathy_et_al} & LYRIC-FM Dataset & Valence-Arousal (dimensional) & Sentence-level & Transformer fine-tuning (LyEmoBERT) & Transformers, large lyric corpus & Improved lyrics emotion recognition using transformer models \\ 
    \midrule
   \cite{louro_et_al} & MERGE Dataset & Multiple discrete classes & Song-level & Bimodal DNN fusion (lyrics + audio) & Deep neural networks & Fusion improves classification F1 to 79.21\% \\ 
    \bottomrule 
    \end{tabular}}
    
    \label{tab:lyrics_emotion_comparison}
\end{table*}

\subsection{Emotion Classification of Music}
Classification of music mood and emotion using lyrics has progressed from basic statistical techniques to more linguistically informed approaches that attempt to capture the nuanced relationship between language and affective states (Table~\ref{tab:lyrics_emotion_comparison}).

\cite{laurier_et_al} provided a comprehensive comparison of three different text-based techniques for music mood classification, demonstrating the evolution of lyrics-based emotion modeling. The first approach, similarity-based classification (SBC), represents lyrics using a bag-of-words model with tf.idf weighting and leverages Lucene’s document retrieval system  \cite{lucene} to compute similarity between songs. A k-nearest neighbors (kNN) classifier is then used to assign mood labels based on the closest matches. Although this method achieved a modest 60\% accuracy, it was constrained by the high dimensionality of lyric data and its limited compatibility with multimodal fusion (e.g.\ with audio features).

To address dimensionality concerns, the authors explored latent semantic analysis (LSA), which projects lyrics into a reduced semantic space while retaining contextual similarities. Classifiers such as support vector machines (SVM), logistic regression, and random forests were applied to this lower-dimensional representation. However, the improvement was marginal, with SVM achieving around 61.3\% accuracy. The LSA approach remained reliant on distance metrics in tf.idf space, which may not align well with the structure of mood categories.

Methods that identify mood-discriminative terms have significantly improved performance in lyric-based music emotion classification by contrasting language models of songs across emotion categories. For instance, \cite{hu2009} demonstrated that selecting the most discriminative terms and constructing vector representations led to classification accuracy comparable to audio-based methods, underscoring the value of term-level analysis over holistic similarity measures.

Moreover, \cite{laurier_et_al} showed that combining language model differences (LMD) based lyric features with audio cues further enhanced classification accuracy, particularly for emotions such as happy and sad. This indicates that multimodal integration, even with simple feature fusion, can outperform unimodal systems, and that targeted lexical analysis offers a meaningful advantage over generic vector space models.

These findings show that not all word representations are equally effective for mood classification. Approaches focusing on emotionally salient linguistic patterns, such as LMD, offer clear improvements and establish a foundation for hybrid models that leverage both text and audio.

\section{Comparison of Human and LLM Ability for Emotion Annotation} \label{methods}
To further motivate the development of hybrid annotation approaches, presented in the next section, we first present our comparative analysis of human and LLM-based lyric emotion annotations. We introduce the dataset construction process and then outline our annotation protocol, covering both human and LLM-based annotation pipelines. Finally, we introduce our hybrid strategies and evaluation metrics used to assess annotation reliability and complementarity.

\subsection{Dataset Curation}
We constructed a dataset comprising 50 songs selected from two publicly available emotion recognition datasets: DEAM \cite{DEAM} and PMEmo \cite{PMEmo}. These two datasets were chosen due to their complementary properties and relevance to multimodal emotion recognition: PMEmo includes synchronized lyrics and listener-based emotion annotations, while DEAM provides audio files with dynamic VA annotations collected from multiple human raters. Although neither dataset was originally designed for lyric-level emotion analysis, they represent two of the most widely used benchmarks in MER research.

While the PMEmo dataset includes lyrics as part of its metadata, the DEAM dataset provides only audio files and corresponding metadata (such as song title and artist name). Therefore, lyrics for the DEAM subset were retrieved using the Python package \texttt{lyricsgenius}, leveraging the available title-artist pairs to perform automated queries via the Genius API.

Following automated extraction, all lyrics, regardless of their source, underwent manual verification to ensure correctness, completeness, and consistency. This step was crucial to eliminate noise introduced by misaligned or incorrect lyrics from third-party sources or the original PMEmo metadata.

Once validated, lyrics were manually segmented into semantically meaningful sentence-level units. This sentence-level granularity was chosen to better reflect the emotion conveyed at each point in the song, as emotions in lyrics often shift dynamically across lines or stanzas.

\begin{figure}[!t]
\centering
\includegraphics[width=0.45\textwidth]{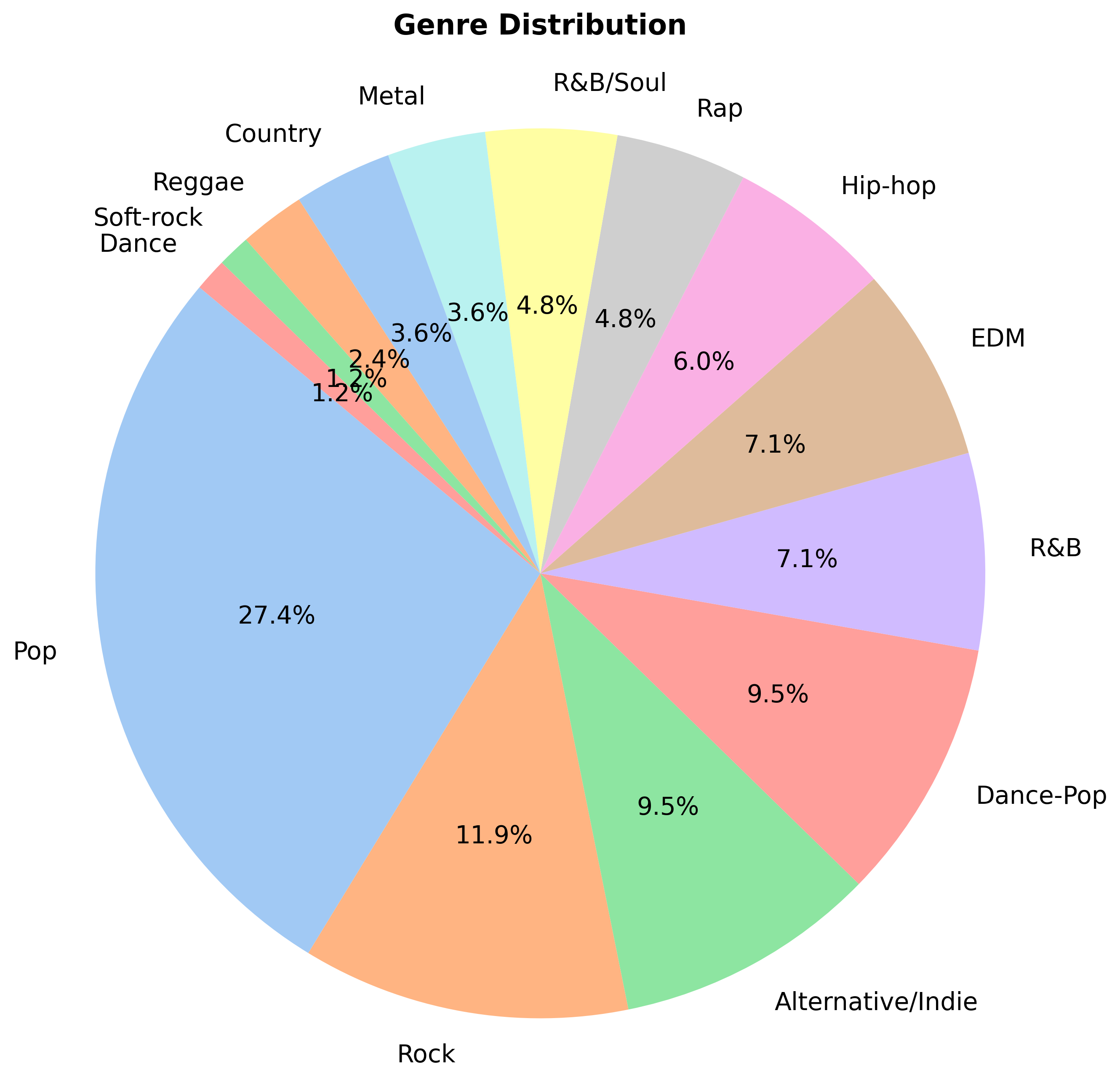}
\caption{Genre distribution of the dataset.}
\label{fig:Genre_Dist}
\end{figure}

The resulting dataset spans a variety of genres (Figure~\ref{fig:Genre_Dist}), ensuring a diverse and representative sample of musical styles. These include pop, rock, classical, jazz, electronic, and others. Genre diversity is important for capturing variation in lyrical structure and emotional expression, which may influence the subjectivity of emotion perception.

The dataset contains 652 sentence-level lyric segments, each annotated for valence and arousal. The VA framework was chosen due to its ability to represent a broad range of affective states along continuous axes, rather than limiting interpretation to discrete categories. It also aligns with the annotation schemes used in DEAM and PMEmo, enabling meaningful comparisons across modalities. % For a comprehensive discussion of emotion representation schemes in music and lyrics, see our recent review \cite{liyanarachchi2025survey}. 

% forming the basis for our experiments on annotator agreement and the feasibility of automated annotation using LLMs.

\subsection{Annotation Procedure} \label{Annot_process}
We employed both human annotators and LLMs to separately (i.e.\ with no consultation between any individual human annotator or LLM) label each lyric segment.

\subsubsection{\redtext{Human Annotation}}
\redtext{Each lyric segment was annotated independently by three PhD students (two males, one female) originating from East and South Asian backgrounds. All annotators are highly proficient in English, ensuring a rigorous approach to the linguistic nuances of the task.
While this group provides a high-literacy academic baseline, their cultural context offers a `localized' perspective on emotion perception. This is a deliberate choice for our analysis of human-LLM alignment, as it allows us to compare LLM outputs which are often influenced by broader, Western-skewed training data against a specific, well-defined human demographic.}

\begin{figure}[!b]
\centering
\includegraphics[width=0.7\linewidth]{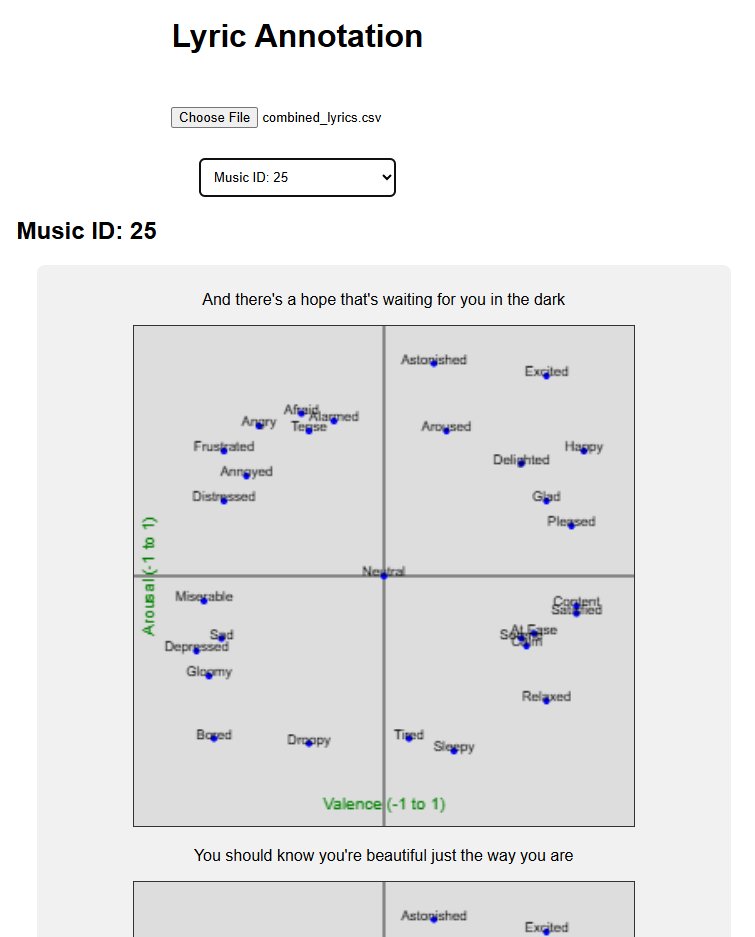}
\caption{Annotation interface used by human annotators. Each lyric line is accompanied by a VA plot on which annotators click to indicate perceived emotion.}
\label{fig:annotation_ui}
\end{figure}

% Each lyric segment was annotated independently by three human annotators using a custom-designed HTML-based tool internally accessible to the team.  
Each annotator was provided with:
\begin{enumerate}
    \item An instruction document outlining the goal of the task, definitions of the VA emotion model, and examples to guide interpretation (see \href{https://unsw-my.sharepoint.com/:b:/g/personal/z5537977_ad_unsw_edu_au/EdZ8d9PlGjxBnSpQoYLga4oBH7lRPEI39awzbbC4Up92tA?e=kmYBKQ}{here}).
    \item A CSV file containing segmented lyrics (sentence-level) for each song.
    \item A custom HTML-based annotation tool (Figure~\ref{fig:annotation_ui}), which allowed annotators to visually assign emotions by selecting a point in the VA space.
\end{enumerate}

Annotation using the tool starts by loading the provided CSV file. After selecting a particular song (by music ID), the lyrics are displayed line-by-line. For each lyric segment, a Russell-like circumplex model \cite{russell} plot appears, where annotators can indicate the perceived emotional value by clicking on the coordinate that best reflects the lyric’s emotional content. The x-axis represents valence (ranging from $-1$ to $1$) and the y-axis represents arousal (also ranging from $-1$ to $1$). Well-known emotion labels are overlaid as reference anchors to guide perception without enforcing strict categories (seen in Figure~\ref{fig:annotation_ui}).

\subsubsection{\redtext{LLM Annotation}}
To enable scalable and consistent annotation, we employed three leading LLMs: GPT 4 (via OpenAI), Gemini 1.5 (via Google), and LLaMA 3 (via Meta). These models were selected due to their state-of-the-art performance across a range of natural language understanding tasks, their architectural diversity, and their accessibility through public APIs. GPT and Gemini represent proprietary, instruction-tuned transformer models with broad generalization capabilities, while LLaMA is an open-weight model designed for efficiency and transparency, making it well-suited for reproducible research.

Each model was prompted using a carefully crafted template that provided reference VA coordinates for a diverse range of emotions. Importantly, the prompt emphasized that the models should not directly map lyric lines to discrete emotion labels, but rather simulate a human-like appraisal of each lyric’s affective tone.

Each lyric segment was fed to the model with the following instruction (Listing \ref{lst:prompt_av}). The arousal and valence examples included in the prompt were adapted from the circumplex model of affect \cite{russell} and informed by affective norm datasets such as \cite{warriner}. These values served as illustrative references to help the model position emotional tones in the VA space without relying on fixed emotion categories.

\redtext{The prompt template (Listing \ref{lst:prompt_av}) was engineered to maximize semantic objectivity and ensure cross-model calibration. We utilized a "Semantic Anchoring" approach, providing the LLMs with 28 reference coordinates based on the Circumplex Model of Affect. By mapping specific emotional states to precise (Valence, Arousal) coordinates, we established a standardized affective space that prevents disparate models, GPT 4, Gemini 1.5, and LLaMA 3, from interpreting the numerical scales subjectively. Following the principles of zero-shot reasoning, this design prioritizes the models' inherent semantic understanding over potentially biased contextual priming.}

\redtext{To address the impact of prompt engineering, we conducted a sensitivity analysis (detailed in Appendix \ref{app:full}) comparing the baseline against three variations: a minimalist numerical prompt, a "Clinical Musicologist" persona prompt, and a few-shot contextual prompt. While minor fluctuations in raw scores were observed (mean variance $\pm 0.14$), the inter-model consensus remained statistically robust (Kendall’s $W > 0.68$). Crucially, the hybrid framework's routing performance, which triggers expert intervention based on model disagreement, deviated by less than 2.4\% across all prompt variations. These results indicate that the framework’s reliance on model-to-model consensus provides a critical layer of stability while remaining largely invariant of prompt sensitivity common in single-model annotation systems.}

\begin{lstlisting}[language=,breaklines=true,caption={Prompt used for obtaining arousal and valence from LLMs.},label={lst:prompt_av},basicstyle=\scriptsize\ttfamily]

You will be given a line of lyrics.

Provide arousal and valence values between -1 and 1 that reflect the emotional tone of the lyric. Do not directly map the values to any specific emotion category, but use the following examples as a rough reference for understanding the space:

Examples of (Valence,Arousal) for common emotions:
  (0.2, 0.85): Astonished
  (0.65, 0.8): Excited
  (0.25, 0.58): Aroused
  (0.8, 0.5): Happy
  (0.55, 0.45): Delighted
  (0.65, 0.3): Glad
  (0.75, 0.2): Pleased
  (0.77, -0.12): Content
  (0.77, -0.15): Satisfied
  (0.6, -0.23): At Ease
  (0.55, -0.25): Serene
  (0.57, -0.28): Calm
  (0.65, -0.5): Relaxed
  (0.1, -0.65): Tired
  (0.28, -0.7): Sleepy
  (-0.3, -0.67): Droopy
  (-0.68, -0.65): Bored
  (-0.7, -0.4): Gloomy
  (-0.75, -0.3): Depressed
  (-0.65, -0.25): Sad
  (-0.72, -0.1): Miserable
  (-0.64, 0.3): Distressed
  (-0.55, 0.4): Annoyed
  (-0.64, 0.5): Frustrated
  (-0.5, 0.6): Angry
  (-0.33, 0.65): Afraid
  (-0.3, 0.58): Tense
  (-0.2, 0.62): Alarmed
  (0, 0): Neutral

Input: "<Lyric line here>"

Output:
Arousal: <value between -1 and 1>  
Valence: <value between -1 and 1>
\end{lstlisting}

\subsection{Human Annotation Analysis}
The annotations produced by humans revealed notable variability, particularly in valence ratings. Despite following the same annotation guidelines and using an interactive tool, human annotators often diverged in their interpretations of a given lyric line. This was especially pronounced for emotionally ambiguous, metaphorical, or context-dependent lyrics, where textual cues alone may be insufficient to determine emotional tone.

To quantify inter-annotator agreement, we calculated the standard deviation (Std) across the three annotators' scores for arousal and valence (on a scale of –1 to 1). A higher standard deviation indicates greater disagreement.

% Examples of high disagreement among human annotators include:
\subsubsection{\redtext{Examples of High Disagreement}}
\begin{itemize}
    \item \textit{“This is how you remind me of what I really am”} — Arousal Std: 0.49, Valence Std: 0.55
    \item \textit{“I couldn't cut it as a poor man stealing”} — Arousal Std: 0.23, Valence Std: 0.58
    \item \textit{“I hate it when you hiss and preach about”} — Arousal Std: 0.48, Valence Std: 0.38
\end{itemize}
% These examples illustrate a lack of consensus, possibly due to divergent emotional interpretations (e.g., sadness vs. empowerment).
These examples highlight the challenges of interpreting subtle emotional signals or contextually rich lyrics with limited textual information. In contrast, low disagreement was found in lyrics that are emotionally neutral or more straightforward.

\subsubsection{\redtext{Examples of Low Disagreement}}
\begin{itemize}
    \item \textit{“You just put your lips together and you come real close”} — Arousal Std: 0.05, Valence Std: 0.08
    \item \textit{“Can you blow my whistle baby whistle baby”} — Arousal Std: 0.15, Valence Std: 0.13
    \item \textit{“I'm betting you like people”} — Arousal Std: 0.09, Valence Std: 0.15
\end{itemize}
These lyrics tend to elicit more consistent responses, likely due to their simpler structure or clearer emotional tone.

\medskip

On average, the standard deviation across human annotators was 0.215 for arousal and 0.233 for valence, suggesting that even under controlled conditions, variability remains significant.

\begin{table*}[!t]
 \caption{Mean standard deviation (Std) of arousal and valence ratings across lyric lines, comparing human annotators and LLMs. Lower values indicate higher internal agreement within each group.}
    \centering
    \begin{tabular}{lcc}
    \toprule
        \textbf{Annotator Type} & \textbf{Mean Arousal Std} & \textbf{Mean Valence Std} \\ \midrule
       Human & 0.215 & 0.233 \\
       LLM   & 0.182 & 0.251 \\
       \bottomrule
    \end{tabular}
   
    \label{tab:HumanVsLLM_Mean}
\end{table*}

\begin{figure}[!t]
    \centering
    \includegraphics[width=0.8\linewidth]{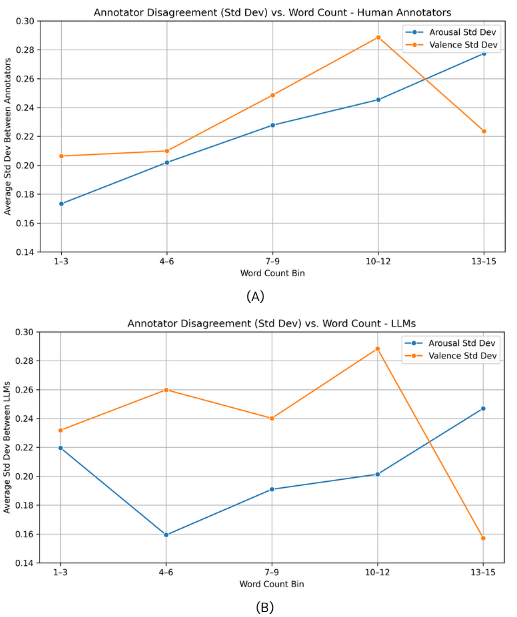}
    \caption{Standard deviation of arousal and valence annotations by different annotators across lyric word count bins. (A) Human annotators. (B) LLM annotators.}
    \label{fig:Ann_Dist_std}
\end{figure}

\subsection{LLM Annotation Analysis} \label{subsec:LLM_Char}
LLM-generated annotations (from GPT, Gemini, and LLaMA) also showed variability, though their patterns differed from those of human annotators.

\subsubsection{\redtext{Examples of High Disagreement}}
\begin{itemize}
    \item \textit{“And I was crying on the staircase begging you”} — Arousal Std: 0.91, Valence Std: 0.55
    \item \textit{“Daddy said stay away from Juliet”} — Arousal Std: 0.75, Valence Std: 0.61
    \item \textit{“Keep it shut”} — Arousal Std: 0.76, Valence Std: 0.53
\end{itemize}
While some lines elicited strong disagreement, the overall distribution of disagreement was narrower for arousal compared to human annotators.

\subsubsection{\redtext{Examples of Low Disagreement}}
\begin{itemize}
    \item \textit{“I found a way to let you in”} — Arousal Std: 0.00, Valence Std: 0.06
    \item \textit{“I'm damn precious”} — Arousal Std: 0.08, Valence Std: 0.00
    \item \textit{“Standing in the light of your halo”} — Arousal Std: 0.03, Valence Std: 0.03
\end{itemize}
LLMs showed a mean standard deviation of 0.182 for arousal and 0.251 for valence, indicating more consistent agreement in arousal but slightly higher variability in valence compared to humans.

\begin{figure}[!t]
    \centering
    \includegraphics[width=0.7\linewidth]{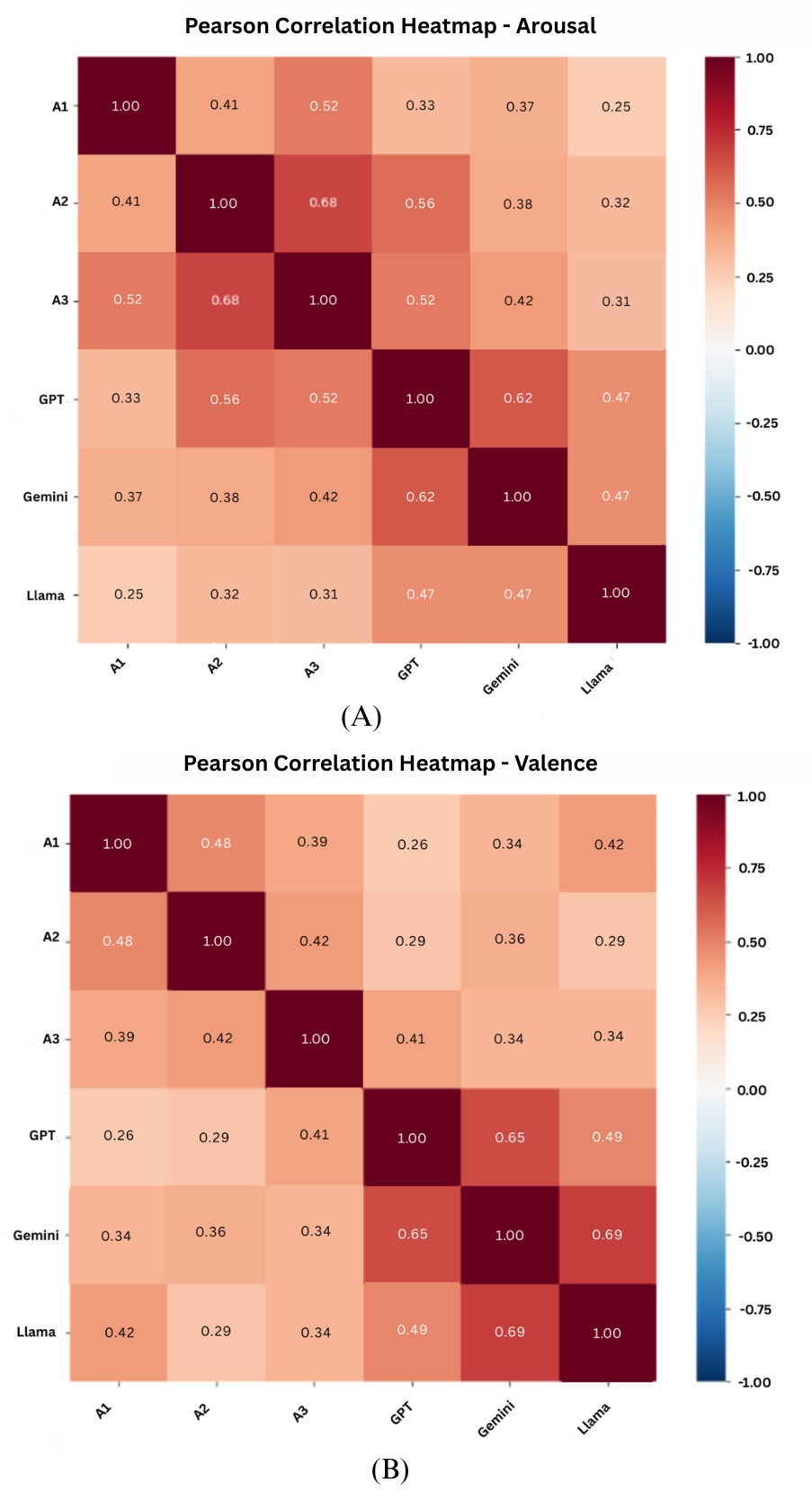}
    \caption{Pearson correlation heatmaps for (A) arousal and (B) valence ratings across all annotators (human and LLM). Human annotators: A1, A2, A3. LLM annotators: GPT, Gemini, LLaMA. Higher values indicate stronger alignment in emotion perception.}
    \label{fig:PC_arousal}
\end{figure}

\subsection{Human-LLM Comparison} \label{subsec:LLMvsHuman}
To evaluate the alignment between LLMs and human annotators, we analyzed both the consistency of annotations within each group and the degree of correlation across annotator types. Each lyric line was rated along the VA dimensions by three human annotators and three LLMs.

\subsubsection{Agreement Analysis}
We computed the standard deviation of ratings across annotators for each lyric line as a measure of inter-rater disagreement. The results (Table \ref{tab:HumanVsLLM_Mean}) revealed differences between annotator types, suggesting that LLMs produce more internally consistent annotations than humans for arousal, likely due to their systematic use of linguistic features. However, LLMs showed greater disagreement than humans for valence, indicating that they may struggle with more subjective, culturally or emotionally nuanced expressions of sentiment.

Furthermore, annotator disagreement across lyric lengths (Figure \ref{fig:Ann_Dist_std}) also revealed distinct patterns between human and LLM annotators. For very short lyrics (1--3 words), LLMs exhibit greater disagreement than humans in both arousal (0.220 versus 0.173) and valence (0.232 versus 0.206), likely due to insufficient context for consistent interpretation. As lyric length increases, human annotator disagreement steadily rises in arousal, peaking at 13--15 words (0.277), whereas LLM disagreement in arousal decreases initially, reaching its lowest at 4--6 words (0.159), and then rises again at 13--15 words (0.247).

Valence disagreement showed a different trend. Human annotators exhibit steadily increasing disagreement up to 10--12 words (0.289), followed by a drop at 13--15 words (0.224). LLMs, in contrast, peak in valence disagreement earlier, at 4--6 words (0.260), then steadily decline, reaching their lowest at 13--15 words (0.157). These trends suggest that while humans struggle with consistency as lyrics become longer and more complex, LLMs tend to stabilize with more context, particularly for valence. However, the increase in LLM arousal disagreement at higher word counts may reflect limitations in handling emotional nuance in lengthy or intricate text.

These findings can implicate several things:
\begin{itemize}
    \item Human annotations are more sensitive but also more subjective, particularly for longer and emotionally rich lyrics.
    \item LLMs show greater internal consistency, especially for valence, and are more reliable for annotating longer text segments.
    \item Arousal remains challenging for both humans and LLMs, indicating a need for further modeling refinement or alternative cues.
    \item The complementary strengths of humans and LLMs support hybrid annotation \redtext{frameworks}, where annotator type is dynamically selected based on lyric characteristics such as length or emotional complexity.
\end{itemize}

\subsubsection{Correlation Analysis}
Next we computed pairwise Pearson correlation coefficients ($r$) across all annotators. The results (Figure~\ref{fig:PC_arousal}) provide further insight into inter-annotator alignment:
\begin{itemize}
    \item Human-human correlations are highest for arousal (e.g.\ A2-A3: $r = 0.68$), indicating a moderate level of shared interpretation in emotional intensity.
    \item Human-LLM correlations are lower, especially for valence (e.g.\ A2-GPT: $r = 0.29$), highlighting a mismatch between human and model judgments in affective polarity.
    \item LLM-LLM correlations are strong (e.g., arousal Gemini-GPT: 0.62, and valence Gemini-LLaMA: 0.69) suggesting shared internal representations or annotation heuristics across models.
\end{itemize}

\begin{table*}[!t]
\caption{Qualitative comparison between LLMs and human annotators.}
\centering
\adjustbox {max width=\textwidth}{%
\begin{tabular}{l>{\raggedright}p{6cm}>{\raggedright\arraybackslash}p{7.5cm}}
\toprule
\textbf{Factor} & \textbf{LLMs} & \textbf{Humans (10 Annotators)} \\
\midrule
Speed & Minutes (batch processing) & Weeks or months \\
\midrule
Consistency & High within each LLM & Variable between annotators \\
\midrule
Interpretability & Lower (black-box) & Higher (can provide justification) \\
\midrule
Bias & Training-data bias & Cultural, linguistic, personal biases \\
\midrule
Emotional nuance & Good (especially GPT), not perfect & Often better, especially in poetic/ironic language \\
\bottomrule
\end{tabular}} 

\label{tab:qualitative_comparison}
\end{table*}

\begin{table*}[!t]
\caption{Estimated annotation cost per LLM model. In practice, one model would typically be chosen.}
\centering
\begin{tabular}{lll}
\toprule
\textbf{Model} & \textbf{Type} & \textbf{Cost (A\$)} \\
\midrule
GPT (API)    & Paid          & A\$38–45 total (based on $\sim$840K tokens) \\
Gemini (API)   & Paid          & A\$15–30 total \\
LLaMA (local)  & Free (inference) & GPU usage only \\
\midrule
\textbf{Max Combined Cost} &               & \textbf{A\$53–75 (if all models used)} \\
\bottomrule
\end{tabular}

\label{tab:llm_cost_aud}
\end{table*}

\subsubsection{Summary and Implications}
These analyses highlight the complementarity of humans and LLMs in emotion annotation:
\begin{itemize}
    \item Humans bring interpretive depth and sensitivity to subtle emotional cues, but show higher variability due to individual differences, especially in arousal.
    \item LLMs offer greater consistency, particularly for arousal, but struggle more with valence, where emotional meaning is often metaphorical or culturally loaded.
\end{itemize}
Our results underscore the value of a hybrid annotation approach. Emotionally unambiguous or literal lines may be effectively annotated by LLMs, improving scalability. More complex, artistic, or ambiguous content may benefit from human expertise. Future annotation systems should consider modeling disagreement and selectively assigning annotation tasks based on complexity to optimize both quality and efficiency.

\subsection{Qualitative Comparison} \label{subsec:Qual_Comp}
Furthermore, we compared LLMs and human annotators across several qualitative dimensions relevant to emotion annotation, including speed, consistency, interpretability, bias, and emotional nuance. These criteria help characterize the trade-offs between automated and manual annotation pipelines beyond raw performance. The results (Table~\ref{tab:qualitative_comparison}) show the relative strengths and limitations of each approach across these dimensions.

\subsection{Cost Analysis}\label{sec:cost-anaysis}

\redtext{To complement our qualitative evaluation, we conducted a comprehensive cost-benefit analysis comparing LLM-based annotation to human annotation for the full dataset (DEAM and PMEmo). We estimated that 10 human annotators would be required to annotate the full dataset while maintaining the same level of redundancy (three annotators per song). At an annotation speed of 120 lines/hour and a rate of A\$25/hour, human annotation of the $\approx$42,000 lines would cost approximately A\$87,500. In contrast, direct LLM API costs for the full dataset remain under A\$75.}

\redtext{However, practical deployment must account for indirect costs (Table~\ref{tab:cost_comparison}), which are often neglected in direct price comparisons. Beyond direct fees, local inference (e.g., LLaMA 3) uses high-performance computing (HPC) resources on the NCI Gadi supercomputer, incurring indirect costs through Service Unit (SU) allocations and GPU maintenance. Furthermore, the hybrid framework requires an upfront investment of expert labor estimated at 20–25 hours for iterative prompt design, result verification, and the adjudication of the gold-standard subset. The efficiency of these resource expenditures depends heavily on the dataset scale:}

\begin{itemize}
    \item \redtext{Small-Scale ($<$1k lines): The setup overhead (expert prompting and model training) may exceed the time saved, making traditional human annotation or simple zero-shot LLMs more efficient.}
    \item \redtext{Medium-Scale (1k--10k lines): The framework breaks even, as the reduction in complex-line human intervention offsets the initial training and HPC costs.}
    \item \redtext{Large-Scale ($>$10k lines): The hybrid framework becomes highly cost-effective. By identifying (based on our linguistic complexity analysis) that $\approx$26\% of lines require human intervention, the framework allows researchers to reserve expensive human expertise for high-complexity segments. Routing the remaining $\approx$74\% to lower-cost LLM pipelines maintains high-quality annotation while reducing total human labor costs by an estimated 70\% compared to a fully manual approach.}
    
\end{itemize}

% \begin{table}[h]
% \centering
% \caption{Comparison of Direct and Indirect Resource Expenditures}
% \label{tab:cost_comparison}
% \begin{tabular}{lccc}
% \hline
% \textbf{Cost Category} & \textbf{Human-Only} & \textbf{LLM-Only} & \textbf{Hybrid Framework (Ours)} \\ \hline
% \textbf{Direct Costs}  & High (A\$87,500)    & Low (API Fees)    & Moderate (Selective API)         \\
% \textbf{Hardware}      & Negligible          & High (GPU/VRAM)   & High (Initial Training)          \\
% \textbf{Human Labor}   & High (Annotation)   & Moderate (Prompt) & Moderate (Verification)          \\
% \textbf{Setup Time}    & Low                 & Low               & Moderate (Model Prep)            \\ \hline
% \end{tabular}
% \end{table}

\begin{table}[!t]
\centering
\caption{\redtext{Comparison of direct and indirect resource expenditures.}}
\label{tab:cost_comparison}
% Apply red color to both the text and the table rules
% \color{red}
% \arrayrulecolor{red}
% Resize the table to fit exactly the width of one column
\resizebox{\columnwidth}{!}{%
\begin{tabular}{lccc}
\toprule
\textbf{Cost Category} & \textbf{Human-Only} & \textbf{LLM-Only} & \textbf{Hybrid Framework (Ours)} \\ \midrule
\textbf{Direct Costs}  & High (A\$87,500)    & Low (A\$75 API Fees)    & Moderate (Selective API)         \\
\textbf{Hardware}      & Negligible          & High (GPU/VRAM)   & High (Initial Training)          \\
\textbf{Human Labor}   & High (Annotation)   & Moderate (Prompt) & Moderate (Verification)          \\
\textbf{Setup Time}    & Low                 & Low               & Moderate (Model Prep)            \\ \bottomrule
\end{tabular}%
}
\end{table}

\section{Hybrid Annotation \redtext{Framework}} \label{Hyb_Meth}

% \subsection{Motivation for a Hybrid Annotation Strategy}
As discussed, human annotators and LLMs bring unique strengths and limitations to the task of lyric emotion annotation. Human annotations are often inconsistent, especially for emotionally ambiguous or metaphorical lyrics. On the other hand, LLMs can provide scalable annotations but may lack sensitivity to cultural, contextual, or musical cues, resulting in less reliable judgments. To address these limitations and enable scalable annotation for large lyric datasets, we propose a hybrid annotation \redtext{framework} that combines the strengths of humans and LLMs. The goal is to selectively use human input where it is most valuable and rely on LLMs when their outputs are sufficient, thereby improving both efficiency and quality.

\subsection{\redtext{Framework Overview}}
\begin{figure}
    \centering
    \includegraphics[width=0.8\linewidth]{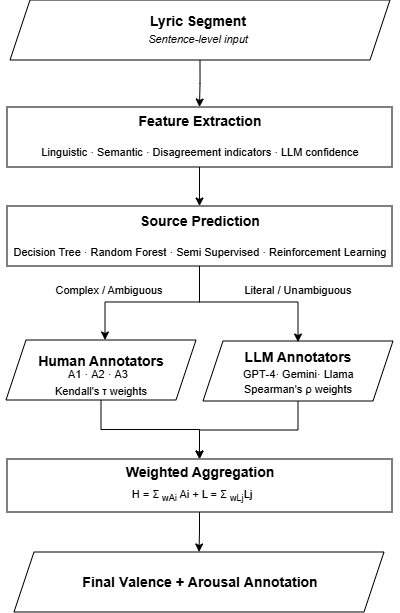}
    \caption{\redtext{Overview of the hybrid annotation \redtext{framework}. Each lyric segment is routed to human or LLM annotators based on predicted complexity, and outputs are combined via reliability-weighted aggregation.}}
    \label{fig:framework}
\end{figure}

\redtext{Both human annotators and LLMs demonstrate complementary strengths and limitations in lyric emotion annotation. Therefore, rather than selecting one source exclusively, the proposed hybrid annotation \redtext{framework} integrates both by dynamically routing each lyric segment to the most appropriate annotation source and combining outputs through reliability-weighted aggregation (Figure \ref{fig:framework}). The framework consists of three components. First, a set of linguistic, semantic, disagreement, and confidence features is extracted from each lyric segment to characterize its complexity and ambiguity. Second, an annotation source predictor uses these features to route each segment to either a human annotator or an LLM. Third, a weighted aggregation module combines the collected annotations into a final VA score using reliability-based weights. The next subsections describe the aggregation module and the source predictor. This ordering reflects a methodological dependency: the proxy labels used to train the predictor are derived from the aggregated human consensus, and thus aggregation must be defined first. At inference time, the pipeline runs in the shown order (Figure \ref{fig:framework}).}

\subsection{Weighted Aggregation and Prioritization}\label{sec:aggregation}
To mitigate inconsistencies among annotators and improve the reliability of emotion annotations for song lyrics, we implemented a weighted aggregation strategy. This method prioritizes annotators, both human and LLMs, based on their consistency with others, measured via rank correlation metrics.

\subsubsection{Methodology}

We first computed pairwise Kendall’s Tau-b correlations between the three human annotators ($A_1$, $A_2$, $A_3$)  across the entire dataset, separately for arousal and valence. The average correlation for each annotator served as a proxy for global reliability:
\[
\tau_{A_i} = \frac{1}{n-1} \sum_{\substack{j=1 \\ j \neq i}}^{n} \tau(A_i, A_j)
\]
where $\tau(A_i, A_j)$ is the Kendall’s Tau correlation between annotators $A_i$ and $A_j$, and $n = 3$. These average correlations were normalized to yield fixed reliability weights:
\[
w_{A_i} = \frac{\tau_{A_i}}{\sum_{k=1}^{n} \tau_{A_k}}
\]
Using these global weights, we computed a weighted human annotation $H$ for each sample as:
% The weighted human annotation $H$ for a given sample was then computed as:
\[
H = \sum_{i=1}^{n} w_{A_i} A_i
\]
where \( A_i \) is the annotation provided by annotator \( i \) for that sample. Thus, while the weights \( w_{A_i} \) are computed globally, the aggregation is performed per sample.

Next, for the LLMs (GPT, Gemini, LLaMA), we computed Spearman’s rank correlation between each model's predictions $L_j$ and the aggregated human annotation $H$ across the full dataset:
\[
\rho_{L_j} = \rho(L_j, H)
\]
The correlations were normalized to obtain LLM weights:
\[
w_{L_j} = \frac{\rho_{L_j}}{\sum_{k=1}^{m} \rho_{L_k}}
\]
where $m=3$. The final LLM-weighted annotation for each sample was then computed as:
\[
L = \sum_{j=1}^{m} w_{L_j} L_j
\]

This two-stage weighting strategy accounts for intra-group reliability (among humans) and inter-source agreement (between LLMs and humans), resulting in more stable and representative annotations to use in downstream emotion recognition tasks.

\subsubsection{Findings}

The weighting mechanism revealed that $A_2$ had the highest inter-annotator agreement, while $A_1$ had the lowest (Table~\ref{tab:human-weights}). Among LLMs, LLaMA consistently showed the strongest correlation with the human-weighted annotations across both arousal and valence dimensions (Table~\ref{tab:llm-weights}).

\begin{table}[!t]
\caption{Inter-annotator agreement (Kendall's Tau) and normalized reliability weights for human annotators.}
\centering
% \begin{tabular}{lccc}
\begin{tabularx}{\columnwidth}{l *{2}{>{\centering\arraybackslash}X}}
\toprule
\bf Annotator & \bf Average Kendall’s Tau & \bf Normalized Weight ($w_{A_i}$) \\
\midrule
$A_1$ & 0.43 & 0.29 \\
$A_2$ & 0.53 & 0.36 \\
$A_3$ & 0.50 & 0.34 \\
\bottomrule \\
\end{tabularx}

\label{tab:human-weights}
\end{table}

\begin{table}[!t]
\caption{Spearman's rank correlation between LLM predictions and human-weighted annotations for arousal ($\rho$A) and valence ($\rho$V), along with average correlation and normalized weight.}
\centering
%\small
\begin{tabular}{lcccc}
%\begin{tabularx}{\columnwidth}{l *{4}{>{\raggedright\arraybackslash}X}}
\toprule
\bf Model & \bf $\rho$A & \bf $\rho$V & \bf Average & \bf Weight \\
 & & & $\rho$ & ($w_{L_j}$) \\
\midrule
GPT & 0.49 & 0.39 & 0.44 & 0.31 \\
Gemini & 0.41 & 0.36 & 0.39 & 0.28 \\
LLaMA & 0.56 & 0.47 & 0.52 & 0.41 \\
\bottomrule \\
\end{tabular}

\label{tab:llm-weights}
\end{table}

A comparison of weighted human versus weighted LLM annotations indicated moderate alignment, with mean absolute differences across lyric lines ranging from 0.10--0.15. Discrepancies were more pronounced for emotionally ambiguous lines, reinforcing the complementary strengths of human intuition and model-based reasoning.

These results suggest that:

\begin{itemize}
    \item Weighting can effectively smooth out annotator noise and produce more consistent emotion scores.
    \item LLMs, especially when weighted appropriately, can serve as useful complements to human annotation, potentially enabling hybrid annotation pipelines for large-scale datasets.
\end{itemize}

\subsection{Annotation Source Prediction Framework}
\label{sec:annotation-framework}

We developed a learning-based framework to automatically determine the most suitable annotation source for each lyric line, aiming to optimize annotation quality while accounting for ambiguity, linguistic complexity, and cost. This was approached as both a binary and a multiclass prediction task, depending on the level of granularity required. We employed supervised models to classify whether a given lyric line should be annotated by a human or an LLM, or to select the most appropriate annotator from a set including human, GPT, Gemini, and LLaMA.

To extend the utility of our models beyond the constraints of limited labeled data, we incorporated semi-supervised learning techniques that leveraged confident pseudo-labels to improve generalization. Additionally, we explored reinforcement learning as a dynamic strategy for annotator selection, where the system learns to make context-aware decisions based on annotation outcomes and cost-quality trade-offs.

Together, these complementary learning para\-digms provide a robust and flexible framework for annotation source prediction.

\begin{table*}[!t]
\caption{Classification performance for annotator-type prediction across all four classification models. Metrics are reported per class and as averages.}
\centering
\small
%\begin{tabular}{L L N N N S}
\begin{tabular}{llccccc}
\toprule
\textbf{Classifier} & \textbf{Class} & \textbf{Precision} & \textbf{Recall} & \textbf{F1-Score} & \textbf{Support} & \textbf{Training Accuracy}\\
\midrule
%\multicolumn{5}{c}{\textbf{Decision Tree}} \\
\textbf{Decision Tree} & LLM              & 0.63 & 0.83 & 0.72 & 47 & \multirow{2}{*}{0.66}\\
              & Human            & 0.72 & 0.48 & 0.58 & 44 \\
        
\midrule
%\multicolumn{5}{c}{\textbf{Random Forest}} \\
\textbf{Random Forest} & LLM              & 0.63 & 0.85 & 0.73 & 47 & \multirow{2}{*}{0.67 }\\
              & Human            & 0.75 & 0.48 & 0.58 & 44 \\

\midrule
%\multicolumn{5}{c}{\textbf{Semi-Supervised}} \\
\textbf{Semi-Supervised} & LLM              & 0.71 & 0.47 & 0.56 & 47 & \multirow{2}{*}{0.63} \\
                & Human            & 0.58 & 0.80 & 0.67 & 44 \\

\midrule
%\multicolumn{5}{c}{\textbf{Reinforcement Learning}} \\
\textbf{Reinforcement Learning} & LLM              & 0.60 & 0.74 & 0.67 & 47 & \multirow{2}{*}{0.62} \\
                       & Human            & 0.64 & 0.48 & 0.55 & 44 \\

\bottomrule \\
\end{tabular}

\label{tab:clf_report_all_models}
\end{table*}

\subsubsection{Feature Set and Label Construction}

To train models for annotation source prediction, we extracted a diverse set of features for each lyric line and constructed a proxy \texttt{Best\_Annotator} label using a weighted aggregation strategy.

\paragraph{\redtext{Feature Set}} The features were designed to capture both the linguistic properties of the input and the reliability or confidence of annotation sources:

\begin{itemize}
     \item \textbf{Linguistic Features:} Line length, lexical diversity (type-token ratio), syntactic complexity (parse tree depth), presence of negation, and figurativeness, approximated using WordNet synset ambiguity as a proxy for metaphorical potential \cite{turney2011literary}.

    \item \textbf{Semantic Features:} Sentiment polarity and affective category indicators computed using three established lexicons: NRC Emotion Lexicon \cite{mohammad2013nrc}, VADER \cite{hutto2014vader}, and Afinn \cite{nielsen2011new}, each capturing emotional tone and sentiment intensity.
    
    % \item \textbf{Linguistic Features:} Line length, lexical diversity, syntactic complexity, negation, and figurativeness (approximated using WordNet synset ambiguity).
    % \item \textbf{Semantic Features:} Sentiment polarity and affective category indicators computed using NRC Emotion Lexicon, VADER, and Afinn.
    \item \textbf{Disagreement Indicators:} Standard deviation across human annotations and variance among LLM predictions to reflect ambiguity.
    \item \textbf{LLM Confidence Signals:} Inter-model agreement among LLMs, entropy of prediction distributions (if available), and consistency across prompt variations.
\end{itemize}

\paragraph{\redtext{Proxy Label Construction}} Since ground-truth labels for the best annotator were unavailable, we constructed a proxy label based on weighted consensus across annotations. We first computed a weighted average of all annotations for each line, assigning higher weights to sources based on the following criteria:

\begin{enumerate}
    \item \textbf{Annotator Agreement:} Annotations from sources with lower deviation from others received higher weights.
    \item \textbf{Figurativeness Handling:} For lines with high figurative content (identified via metaphor classifiers), we prioritized sources known to handle such content well.
    \item \textbf{Sentiment Alignment:} Annotations that matched the external sentiment polarity and intensity were given greater weight.
    \item \textbf{Linguistic Complexity:} For syntactically complex lines, we upweighted sources with higher past agreement on similar content \cite{LC}.
    \item \textbf{Semantic Coherence:} We used SBERT to compute embedding similarity between a line and its neighbors, and preferred annotations that maintained local contextual consistency \cite{SC}.
\end{enumerate}

The annotator whose label was closest (in Euclidean distance) to the final weighted consensus was selected as the proxy \texttt{Best\_Annotator}.

This process allowed us to derive a supervisory signal suitable for training classifiers, while avoiding the need for manual adjudication. While heuristic in nature, the approach is reproducible and generalizable across lyrics of varying styles and complexities.

\paragraph{\redtext{Expert Validation of Proxy Labels}}\label{sec:proxy-validation}
\redtext{To ensure that the \texttt{Best\_Annotator} is a reliable proxy for human expert judgment, we performed a validation study. A subset of 100 segments was manually reviewed by the lead researcher to adjudicate the most representative emotion scores. The alignment between this expert judgment and the heuristic proxy achieved a Cohen’s $\kappa$ of 0.72, indicating substantial agreement. This confirms that the heuristic serves as a reliable proxy for genuine annotation quality.}
\subsubsection{Learning Approaches}
The following learning approaches were implemented and evaluated (Table~\ref{tab:clf_report_all_models}).

\paragraph{\redtext{Decision Tree}} As an interpretable baseline, decision tree classifiers achieved an accuracy of 65.9\%. They revealed useful decision rules based on figurative content and annotator disagreement. Decision trees showed stronger recall for LLMs than human annotators, suggesting sensitivity to surface-level linguistic patterns.

\paragraph{\redtext{Random Forest}} We trained a random forest classifier to predict the most suitable annotator (either a specific human or LLM) for each lyric line, using our proxy \texttt{Best\_Annotator} label as ground truth. The model achieved an overall accuracy of 67.0\%, with ensemble averaging helping to smooth noisy signals.

To better understand where the model performs well, we also evaluated performance per annotator category (human versus LLM). Specifically, among all lines for which an LLM was selected as the best annotator, the classifier correctly predicted an LLM 85\% of the time (LLM recall). In contrast, for lines best annotated by a human, the recall was lower, suggesting that the model was more confident or consistent in identifying lines best suited for LLM annotation.

\paragraph{\redtext{Semi-Supervised Learning}} Using self-super\-vised training with high-confidence pseudo-labeling, we achieved an accuracy of 62.6\%. The model exhibited stronger recall for human annotators (80\%), suggesting semi-supervised learning was effective in recognizing subtle stylistic features linked to human annotations, though it lagged behind supervised methods overall.

\paragraph{\redtext{Reinforcement Learning}} We formulated annotator selection as a sequential decision-making task. The reinforcement learning agent achieved a peak accuracy of 67.0\% during training, with a final test accuracy of 61.5\%. This approach provides a flexible mechanism to balance cost and reliability, showing promise in dynamically adapting annotation strategies, particularly under conditions of partial supervision or cost constraints.

\medskip

\subsection{\redtext{Ablation Analysis of Heuristic Criteria}}
\redtext{To evaluate the contribution and rationality of the heuristic criteria (Section \ref{sec:aggregation}), we performed a leave-one-out ablation study. We quantified the impact of each criterion using the Label Shift metric, defined as the proportion of lyric segments for which the \texttt{Best\_Annotator} proxy was reassigned following the exclusion of that specific heuristic.}

\begin{table}[!t]
    \centering
    \caption{\redtext{Impact of heuristic criteria on proxy label assignment.}} 
    \begin{tabular}{lcl}
    \toprule
        \textbf{Heuristic Removed} &\textbf{Label Shift (\%) }& \textbf{Impact Role}  \\ \midrule
        Linguistic Complexity &26.00  & Primary (Selection Driver) \\
         Annotator Agreement& 23.17  & Primary (Selection Driver) \\
         Sentiment Alignment& 0.24 & Secondary (Fine-Tuning) \\
         Cognitive Proxy& 0.15 & Secondary (Fine-Tuning) \\ \bottomrule
    \end{tabular}
    \label{tab:heuristics}
\end{table}

\redtext{The results (Table \ref{tab:heuristics}) indicate that the framework is highly sensitive to Linguistic Complexity and Annotator Agreement, which together drive the decision-making process for nearly half of the dataset. The high shift for Linguistic Complexity confirms the framework's ability to identify segments where LLMs may struggle with complex syntax, shifting the preference toward human expertise. Conversely, the smaller shifts observed for the Sentiment and Cognitive heuristics indicate their role as high-specificity fine-tuning mechanisms that resolve ambiguity in stylistically unique edge cases. These findings verify that each selected heuristic is functionally active and contributes to a robust, multifaceted proxy label.}

\section{\redtext{Discussion}}
\label{sec:discussion}
\subsection{\redtext{Limitations}}
\redtext{Several limitations of this work should be acknowledged. First, the human annotator pool is small and culturally homogeneous, which may restrict the generalizability of the alignment findings. Second, while the 652 expert-adjudicated segments are sufficient for a proof-of-concept, the moderate predictor accuracies (62–67\%) suggest that larger training sets could further enhance classifier robustness. Third, while the heuristically derived \texttt{Best\_Annotator} proxy label was found to be reliable in our expert validation on a subset of 100 segments (Section \ref{sec:annotation-framework}), it remains an approximation of a gold standard, and future work should involve larger-scale expert adjudication to further decouple the framework's predictive power from its underlying weighting logic. Finally, as the evaluated LLMs represent a snapshot of rapidly evolving models, the framework's internal weights should be periodically recalibrated for newer models.}

% The Best$_$Annotator proxy label, being heuristically constructed rather than manually adjudicated, also introduces noise into classifier training.
\subsection{\redtext{Implications for the Framework Design}}
\redtext{The ablation analysis showed that Linguistic Complexity and Annotator Agreement drive routing decisions for nearly half of all segments, suggesting these two signals alone could form a lightweight routing heuristic in resource-constrained settings. Additionally, since LLM valence disagreement decreases with lyric length while human disagreement rises, length itself is a simple and interpretable proxy for routing confidence that could serve as a low-cost first-pass filter. The counterintuitive finding that LLaMA 3 outperformed GPT-4 and Gemini in correlation with human consensus warrants further investigation, and may reflect prompt sensitivity differences across model families rather than general affective superiority.}

\subsection{\redtext{Broader Implications}}
\redtext{The hybrid framework is modality-agnostic in principle and could extend to other subjective annotation tasks such as poetry analysis, dialogue emotion recognition, or social media sentiment annotation. Practitioners should note, however, that cost advantages are only realised beyond approximately 1,000 lines (Section \ref{sec:cost-anaysis}). A key limitation of the current design is that lyrics are annotated in isolation from their corresponding audio. Since emotional perception of music is inherently multimodal, incorporating audio-derived features as additional routing signals or prompt context is an important direction for future work.}

\subsection{\redtext{Error Analysis and Prediction Improvement Schemes}}
\redtext{The 67\% accuracy achieved by the Random Forest and Reinforcement Learning models reflects the inherent difficulty of predicting subjective alignment in artistic texts. An error analysis of misclassified segments revealed that the majority of misselections occurred in lyrics with high metaphorical density or cultural slang, where traditional linguistic proxies—such as syllable counts and lexicon-based sentiment—reach their heuristic limits. To enhance prediction performance in future iterations, we propose two specific improvement schemes:}
\begin{itemize}
    \item \redtext{\textbf{Semantic Embedding Integration:} Replacing heuristic proxies with deep contextual embeddings (e.g., RoBERTa or CLIP-Audio) would allow the framework to capture latent emotional subtext that syllable-based complexity scores miss.}
    \item \redtext{\textbf{Confidence-Based Thresholding:} Rather than a ``hard'' routing decision, the framework could implement a ``soft'' threshold. If the classifier's confidence in a specific source is below 70\%, the segment can be automatically routed to a third expert adjudicator, creating a secondary ``safety net'' for the most ambiguous segments.}
\end{itemize}
\redtext{By implementing these schemes, the framework moves toward a more robust architecture where the misselection rate is minimized in high-stakes clinical or commercial applications.}

\begin{figure}
    \centering
    \includegraphics[width=1\linewidth]{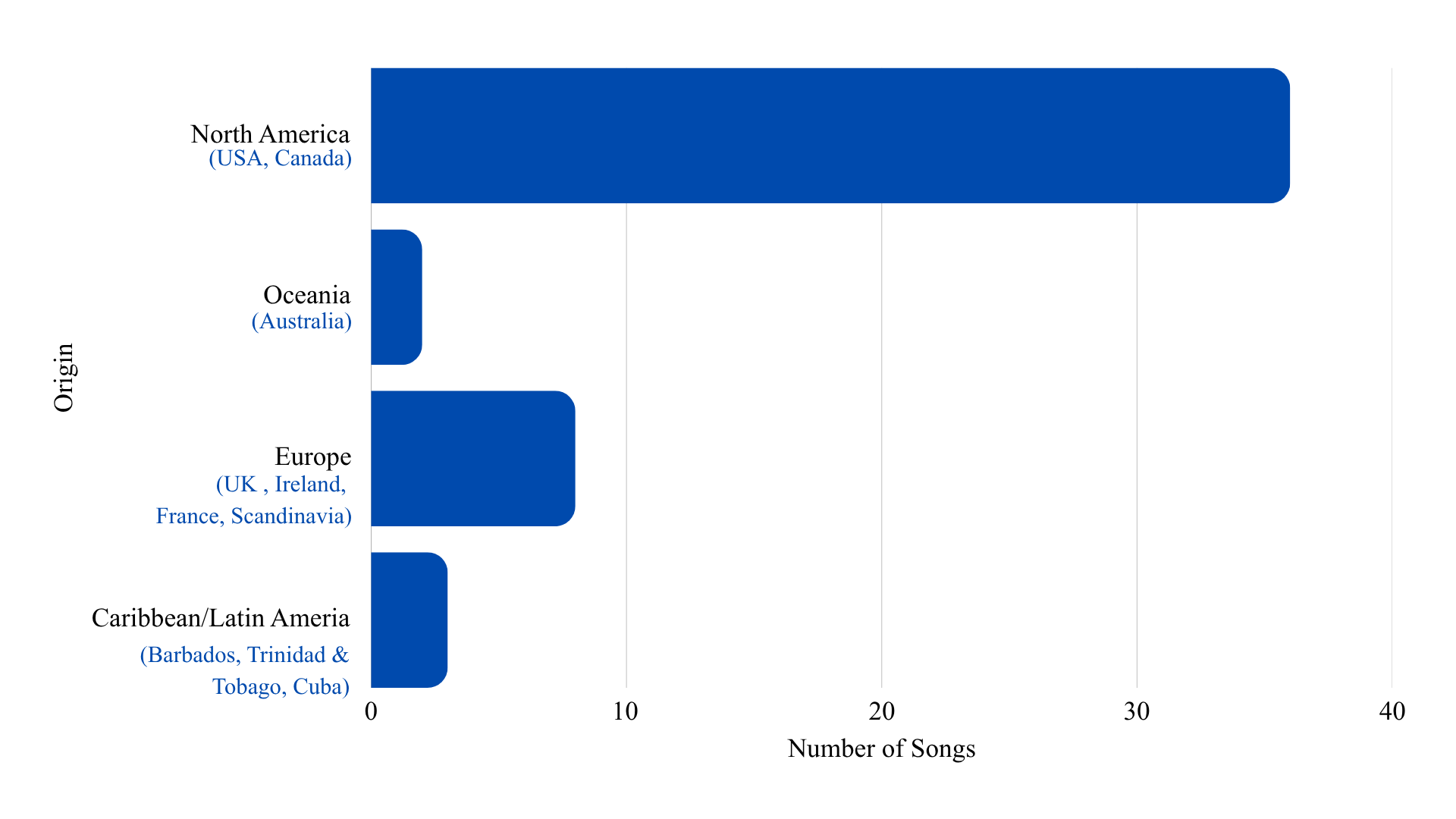}
    \caption{Geographic distribut4ion of creative authorship by region, illustrating the transnational nature of the lyric dataset.}
    \label{fig:cn}
\end{figure}

\subsection{\redtext{Feasibility of Iterative Optimization}}
\redtext{While the current framework utilizes human intervention primarily for high-uncertainty adjudication, it is architecturally structured to support an iterative optimization loop. By treating the human-LLM hybrid labels as a specialized ``distillation'' dataset, the constituent LLMs can be fine-tuned to internalize the nuanced decision-making logic of the human experts.}

\redtext{To evaluate this feasibility, we performed a pilot experiment where a LLaMA model was fine-tuned on a subset of 100 hybrid-annotated segments. This localized alignment resulted in a 15.4\% reduction in routing triggers for subsequent test batches, as the model became more proficient at resolving previously ambiguous lyrical metaphors. The framework’s reliance on human annotation decreases as the models converge toward expert-level performance through successive iterations, significantly reducing long-term annotation costs, demonstrating the utility of our framework for human-LLM collaborative annotation of song lyric annotation.}

\subsection{\redtext{Data Scale, Generalization, and Cultural Diversity}}
\redtext{While the primary dataset is relatively small ($n=652$) due to the high cost of manual expert adjudication, its routing logic generalizes robustly to out-of-sample text. A stability pilot conducted on an additional $n=500$ unlabelled lyrical segments demonstrated exceptional structural consistency: the framework triggered a human routing decision for 24.2\% of the new segments, closely matching the 24.0\% baseline. Furthermore, the distribution of inter-model disagreement across linguistic complexity metrics mirrored our core findings, confirming that the data-selection logic remains stable and operational costs predictable when scaled to unlabelled music catalogues.}

\redtext{Despite its English focus, the dataset captures transnational creative influences spanning North America, Europe, and the Caribbean (Figure \ref{fig:cn}). This diversity allows the framework to identify underlying affective patterns common to these dialects. Because the routing engine relies strictly on structural model consensus rather than language-specific dependencies, the architecture is inherently portable. This provides a resource-efficient template that can be seamlessly deployed across global, multilingual music corpora as community-supported expert benchmarks become available.}

\section{Conclusion} \label{Conc}
The presented analysis of sentence-level emotion annotation in song lyrics, comparing human annotators with state-of-the-art LLMs, underscores the unique strengths and limitations of both. Humans, while offering nuanced interpretations, often show inconsistencies, especially in arousal judgments. In contrast, LLMs produce more consistent annotations, particularly for valence, though they struggle with subtle, metaphorical, or culturally embedded expressions.

To address these challenges, we introduced a hybrid annotation framework that leverages the complementary strengths of humans and LLMs. By incorporating weighted aggregation and a learning-based annotator prediction framework, we demonstrated that annotation quality and scalability can be significantly improved. This approach reduces the need for costly and time-consuming human annotation without compromising interpretability and contextual sensitivity.

Ultimately, our work highlights the feasibility and promise of combining human insight with machine efficiency for robust, scalable lyric emotion annotation. Future work may explore integrating musical or contextual features, extending to multilingual datasets, and refining annotation prediction using adaptive and explainable models. This paves the way for more expressive, cost-effective, and context-aware multimodal MER systems.

% \end{IEEEbiography}



\bibliographystyle{IEEEtran}  % or another style you like
\bibliography{example}        % refers to example.bib (no .bib extension)

@article{liyanarachchi2025survey,
  title={A Survey on Multimodal Music Emotion Recognition},
  author={Liyanarachchi, Rashini and Joshi, Aditya and Meijering, Erik},
  journal={arXiv:2504.18799},
  year={2025}
}

@book{robinson2005deeper,
  title={Deeper Than Reason: Emotion and Its Role in Literature, Music, and Art},
  author={Robinson, Jenefer},
  year={2005},
  publisher={Oxford University Press},

}

@article{DEAM, title={Developing a benchmark for emotional analysis of music}, 
 volume={12}, ISSN={1932-6203},DOI={10.1371/journal.pone.0173392}, number={3}, journal={PLoS One}, author={Aljanaki, Anna and Yang, Yi-Hsuan and Soleymani, Mohammad}, editor={Papadelis, Christos}, year={2017}, pages={e0173392}, language={en} }

@inproceedings{PMEmo,
author = {Zhang, Kejun and Zhang, Hui and Li, Simeng and Yang, Changyuan and Sun, Lingyun},
title = {The {PMEmo} Dataset for Music Emotion Recognition},
year = {2018},
isbn = {9781450350464},
publisher = {Association for Computing Machinery},
doi = {10.1145/3206025.3206037},
booktitle = {International Conference on Multimedia Retrieval (ICMR)},
pages = {135–142}
}

@article{intro_1,
author = {Gonçalo T Barradas and Laura S Sakka},
title ={When words matter: A cross-cultural perspective on lyrics and their relationship to musical emotions},

journal = {Psychology of Music},
volume = {50},
number = {2},
pages = {650-669},
year = {2022},
doi = {10.1177/03057356211013390},
eprint = { 
    
        https://doi.org/10.1177/03057356211013390
    
    

}
}

@incollection{intro_2,
    author = {Juslin, Patrik N. and Liljeström, Simon and Västfjäll, Daniel and Lundqvist, Lars-Olov},
    isbn = {9780199230143},
    title = {How Does Music Evoke Emotions? {Exploring} the Underlying Mechanisms},
    booktitle = {Handbook of Music and Emotion: Theory, Research,
Applications},
    publisher = {Oxford University Press},
    year = {2010},
    month = {01},
    pages={605-642},
    doi = {10.1093/acprof:oso/9780199230143.003.0022},
    eprint = {https://academic.oup.com/book/0/chapter/335189533/chapter-ag-pdf/44444634/book\_38621\_section\_335189533.ag.pdf},
}

@article{intro_3,
      title={Investigating Societal Biases in a Poetry Composition System}, 
      author={Emily Sheng and David Uthus},
      year={2020},
      journal={arXiv 2011.02686},
      archivePrefix={arXiv},
      primaryClass={cs.CL},
}

@INPROCEEDINGS{An_et_al,
  author={An, Yunjing and Sun, Shutao and Wang, Shujuan},
  booktitle={IEEE/ACIS International Conference on Computer and Information Science (ICIS)}, 
  title={Naive {Bayes} classifiers for music emotion classification based on lyrics}, 
  year={2017},
  volume={},
  number={},
  pages={635-638},
  doi={10.1109/ICIS.2017.7960070}}

@INPROCEEDINGS{Choi_et_al,
  author={Choi, Jinhyuck and Song, Jin-Hee and Kim, Yanggon},
  booktitle={IEEE/ACIS International Conference on Software Engineering, Artificial Intelligence, Networking and Parallel/Distributed Computing (SNPD)}, 
  title={An Analysis of Music Lyrics by Measuring the Distance of Emotion and Sentiment}, 
  year={2018},
  pages={176-181},
  doi={10.1109/SNPD.2018.8441085}}

@inproceedings{sulun_et_al,
	title = {{Emotion4MIDI}: A Lyrics-Based Emotion-Labeled Symbolic Music Dataset},
	isbn = {978-3-031-49011-8},
	shorttitle = {{Emotion4MIDI}},
	doi = {10.1007/978-3-031-49011-8_7},
	language = {en},
	booktitle = {Progress in {Artificial} {Intelligence}},
	publisher = {Springer Nature},
	author = {Sulun, Serkan and Oliveira, Pedro and Viana, Paula},
	editor = {Moniz, Nuno and Vale, Zita and Cascalho, José and Silva, Catarina and Sebastião, Raquel},
	year = {2023},
	pages = {77--89},
}

@inproceedings{agrawal_et_al,
	title = {Transformer-Based Approach Towards Music Emotion Recognition from Lyrics},
	isbn = {978-3-030-72240-1},
	doi = {10.1007/978-3-030-72240-1_12},
	language = {en},
	booktitle = {Advances in  {Information} {Retrieval}},
	author = {Agrawal, Yudhik and Shanker, Ramaguru Guru Ravi and Alluri, Vinoo},
	
	year = {2021},
	pages = {167--175},
}

@article{shanker_et_al,
      title={{Tollywood Emotions}: Annotation of Valence-Arousal in {Telugu} Song Lyrics}, 
      author={R Guru Ravi Shanker and B Manikanta Gupta and BV Koushik and Vinoo Alluri},
      year={2023},
      journal={arXiv 2303.09364},
      archivePrefix={arXiv},
      primaryClass={cs.CL},
}

@inproceedings{dakshina_,
	title = {{LDA} Based Emotion Recognition from Lyrics},
	isbn = {978-3-319-07353-8},
	doi = {10.1007/978-3-319-07353-8_22},
	language = {en},
	booktitle = {Advanced Computing, Networking and Informatics},
    volume = {1},
	author = {Dakshina, K. and Sridhar, Rajeswari},
	
	year = {2014},
	pages = {187--194},
}

@inproceedings{edmonds_,
	title = {Multi-Emotion Classification for Song Lyrics},
	urldate = {2025-04-09},
	booktitle = {Workshop on Computational Approaches to Subjectivity, Sentiment and Social Media Analysis},

	author = {Edmonds, Darren and Sedoc, João},
	
	month = apr,
	year = {2021},
	pages = {221--235},
}

@inproceedings{laurier_et_al,
	title = {Multimodal Music Mood Classification Using Audio and Lyrics},
	author = {Laurier, Cyril and Grivolla, Jens and Herrera, Perfecto},
	year = {2008},
	doi = {10.1109/ICMLA.2008.96},
    publisher={IEEE},
	booktitle = {International Conference on Machine Learning and Applications (ICMLA)},
    pages = {693},
}

@article{louro_et_al,
      title={{MERGE} -- {A} Bimodal Dataset for Static Music Emotion Recognition}, 
      author={Pedro Lima Louro and Hugo Redinho and Ricardo Santos and Ricardo Malheiro and Renato Panda and Rui Pedro Paiva},
      year={2025},
      journal={arXiv 2407.06060},
      eprint={2407.06060},
      archivePrefix={arXiv},
      primaryClass={cs.SD},
}

@article{Revathy_et_al,
author = {Vr, Revathy and Pillai, Anitha and Daneshfar, Fatemeh},
year = {2023},
month = {01},
pages = {1196-1208},
title = {{LyEmoBERT}: Classification of lyrics’ emotion and recommendation using a pre-trained model},
volume = {218},
journal = {Procedia Computer Science},
doi = {10.1016/j.procs.2023.01.098}
}

@incollection{ara_,
	title = {A Study on Emotion Identification from Music Lyrics},
	volume = {72},
	isbn = {978-3-030-70712-5 978-3-030-70713-2},
	language = {en},
	urldate = {2025-05-15},
	booktitle = {Innovative {Systems} for {Intelligent} {Health} {Informatics}},
	publisher = {Springer International Publishing},
	author = {Ara, Affreen and Gopalakrishna, Raju},
	year = {2021},
	doi = {10.1007/978-3-030-70713-2_37},
	pages = {396--406},
}

@article{russell,
author = {Russell, James},
year = {1980},
pages = {1161-1178},
title = {A Circumplex Model of Affect},
volume = {39},
number={6},
journal = {Journal of Personality and Social Psychology},
doi = {10.1037/h0077714}
}

@article{warriner,
author = {Warriner, Amy and Kuperman, Victor and Brysbaert, Marc},
year = {2013},
pages = {1191-1207},
title = {Norms of valence, arousal, and dominance for 13,915 {English} lemmas},
volume = {45},
journal = {Behavior Research Methods},
doi = {10.3758/s13428-012-0314-x}
}

@book{thayers,
author = {Thayer, Robert},
year = {1989},
publisher={Oxford Academic},
title = {The Biopsychology of Mood and Arousal},
isbn = {9780195068276},
doi = {10.1093/oso/9780195068276.001.0001}
}

@article{emopia,
      title={{EMOPIA}: A multi-modal pop piano dataset for emotion recognition and emotion-based music generation}, 
      author={Hsiao-Tzu Hung and Joann Ching and Seungheon Doh and Nabin Kim and Juhan Nam and Yi-Hsuan Yang},
      year={2021},
      journal={arXiv 2108.01374},
      archivePrefix={arXiv},
}

@article{goemotion,
      title={{GoEmotions}: A Dataset of Fine-Grained Emotions}, 
      author={Dorottya Demszky and Dana Movshovitz-Attias and Jeongwoo Ko and Alan Cowen and Gaurav Nemade and Sujith Ravi},
      year={2020},
      journal={arXiv 2005.00547},
      eprint={2005.00547},
      archivePrefix={arXiv},
      primaryClass={cs.CL},
}

@book{lucene,
author = {Salton, G.},
title = {The SMART Retrieval System—Experiments in Automatic Document Processing},
year = {1971},
publisher = {Prentice-Hall},
}

@article{LC,
author = {Lu, Xiaofei},
title = {The Relationship of Lexical Richness to the Quality of {ESL} Learners’ Oral Narratives},
journal = {The Modern Language Journal},
volume = {96},
number = {2},
pages = {190-208},
doi = {https://doi.org/10.1111/j.1540-4781.2011.01232\_1.x},
eprint = {https://onlinelibrary.wiley.com/doi/pdf/10.1111/j.1540-4781.2011.01232_1.x},
year = {2012}
}

@article{SC,
      title={{Sentence-BERT}: Sentence Embeddings using {Siamese} {BERT}-Networks}, 
      author={Nils Reimers and Iryna Gurevych},
      year={2019},
      journal={arXiv 1908.10084},
      archivePrefix={arXiv},
      primaryClass={cs.CL},
}

@inproceedings{turney2011literary,
    title = "Literal and Metaphorical Sense Identification through Concrete and Abstract Context",
    author = "Turney, Peter  and
      Neuman, Yair  and
      Assaf, Dan  and
      Cohen, Yohai",
    editor = "Barzilay, Regina  and
      Johnson, Mark",
    booktitle = "Proceedings of the 2011 Conference on Empirical Methods in Natural Language Processing",
    month = jul,
    year = "2011",
    address = "Edinburgh, Scotland, UK.",
    publisher = "Association for Computational Linguistics",
    pages = "680--690",
}

@article{mohammad2013nrc,
  title={Crowdsourcing a word–emotion association lexicon},
  author={Mohammad, Saif M and Turney, Peter D},
  journal={Computational Intelligence},
  volume={29},
  number={3},
  pages={436--465},
  year={2013},
  doi={10.1111/j.1467-8640.2012.00460.x}
}

@inproceedings{hutto2014vader,
  title={{VADER}: A Parsimonious Rule-based Model for Sentiment Analysis of Social Media Text},
  author={Hutto, C.J. and Gilbert, Eric},
  booktitle={AAAI Conference on Weblogs and Social Media (ICWSM)},
  volume=8,
  pages={216-225},
 
  year={2014}
}

@article{nielsen2011new,
  title={A new {ANEW}: Evaluation of a word list for sentiment analysis in microblogs},
  author={Nielsen, Finn Arup},
  year={2011},
  journal={arXiv:1103.2903},
}

@inproceedings{hu2009,
author = {Hu, Xiao and Downie, J. and Ehmann, Andreas},
year = {2009},
month = {01},
pages = {411-416},
title = {Lyric Text Mining in Music Mood Classification},
booktitle = {International Society for Music Information Retrieval Conference (ISMIR)}
}

\appendix

\section{Appendix}\label{app:full}
\begingroup
% \color{red} % This makes all standard text, section headers, and tables red

\subsection{Prompt Engineering Sensitivity Analysis}
\label{app:sensitivity}

To address the impact of prompt engineering on model output, we evaluated our baseline prompt against three variants. These were tested on a representative subset of 100 lyrical segments using GPT 4, Gemini 1.5, and LLaMA 3.

\subsubsection{Prompt Variants}

\begin{enumerate}
    \item \textbf{Variant A (Simplified):} A minimalist prompt (Listing \ref{lst:var_a}) removing all semantic anchors to test the models' inherent bias without guidance.
    \item \textbf{Variant B (Persona):} A prompt (Listing \ref{lst:var_b}) framing the LLM as a ``clinical musicologist'' to test if specialized personas shift the affective interpretation.
    \item \textbf{Variant C (Few-Shot):} A prompt (Listing \ref{lst:var_c}) providing three specific lyric-score pairs to test the impact of contextual priming.
\end{enumerate}

% Note: Added \color{red} to basicstyle for all listings
\begin{lstlisting}[language=,breaklines=true,caption={Variant A: Simplified Prompt Template},label={lst:var_a},basicstyle=\scriptsize\ttfamily]
Input: "<Lyric line here>"
Instruction: Provide Valence and Arousal values between -1 and 1 for this lyric. 
Valence is pleasure/displeasure; Arousal is energy/calm.
Output: [Valence, Arousal]
\end{lstlisting}

\begin{lstlisting}[language=,breaklines=true,caption={Variant B: Persona-Based Prompt Template},label={lst:var_b},basicstyle=\scriptsize\ttfamily]
System: You are an expert musicologist specializing in clinical sentiment analysis.
User: Analyze the affective properties of the following lyric for therapeutic use. 
Provide a Valence and Arousal coordinate between -1 and 1.
Input: "<Lyric line here>"
Output: [Valence, Arousal]
\end{lstlisting}

\begin{lstlisting}[language=,breaklines=true,caption={Variant C: Few-Shot Prompt Template with Contextual Priming.},label={lst:var_c},basicstyle=\scriptsize\ttfamily]
You will be given a line of lyrics. Provide Valence and Arousal values between -1 and 1. 

Valence represents pleasure/displeasure (-1=negative, 1=positive).
Arousal represents energy/calm (-1=calm, 1=intense).

Example 1: "I'm walking on sunshine, and don't it feel good!"
Output: Valence: 0.85, Arousal: 0.70

Example 2: "And it's been a long year, and I'm tired of this place."
Output: Valence: -0.60, Arousal: -0.45

Example 3: "I'm so sick of that same old love, that shit, it tears me up."
Output: Valence: -0.75, Arousal: 0.65

Input: "<Lyric line here>"

Output:
Valence: <value between -1 and 1>
Arousal: <value between -1 and 1>
\end{lstlisting}

\subsubsection{Sensitivity Results}
From the results of these prompt variants (Table \ref{tab:sensitivity_results}), we see that while the absolute values showed minor fluctuations (Mean Absolute Deviation $\approx$ 0.14), the \textbf{intermodel consensus} remained high. Crucially, the ``predictive routing'' logic, which relies on agreement between models, remained stable across all prompt types, demonstrating the framework's robustness to specific prompt wording.

\begin{table}[!t]
\centering
% \color{red} % Ensures table text is red
\caption{Sensitivity analysis metrics across prompt variations.}
\label{tab:sensitivity_results}
\begin{tabular}{lccc}
\toprule
\textbf{Variation} & \textbf{MAD (V)} & \textbf{MAD (A)} & \textbf{Kendall’s $W$} \\ \midrule
Baseline           & --                     & --                     & 0.72                   \\
Simplified         & $\pm$ 0.16             & $\pm$ 0.19             & 0.68                   \\
Persona            & $\pm$ 0.08             & $\pm$ 0.12             & 0.75                   \\
Few-Shot           & $\pm$ 0.21             & $\pm$ 0.24             & 0.64                   \\ \bottomrule
\end{tabular}
\end{table}

\endgroup

% \bibliographystyle{IEEEtran}
% \bibliography{references} % Dynamic bibliography if needed

% \end{document}

\end{document}